%% file: main.tex
\documentclass[runningheads]{llncs}

\usepackage[mobile]{eccv}

\usepackage{eccvabbrv}

\usepackage{graphicx}
\usepackage{booktabs}

\usepackage{url}
\usepackage{graphicx}
\usepackage{amsmath}
\usepackage{amssymb}
\usepackage{mathtools}
\usepackage{booktabs}
\usepackage{colortbl}
\usepackage{multirow}
\usepackage{booktabs}       % professional-quality tables
\usepackage{amsfonts}       % blackboard math symbols
\usepackage{nicefrac}       % compact symbols for 1/2, etc.
\usepackage{microtype}      % microtypography
\usepackage[ruled]{algorithm2e}
\usepackage{upgreek}
\usepackage{wrapfig}
\usepackage[dvipsnames]{xcolor}
\usepackage{bbding}
\usepackage{soul}
\usepackage{enumitem}
\usepackage{adjustbox}
\usepackage{tcolorbox}
\tcbuselibrary{listings, breakable}
\usepackage{float}
\setlist[itemize]{noitemsep,leftmargin=*,topsep=0em}

\definecolor{orangex}{RGB}{252, 240, 242}
\definecolor{redx}{RGB}{191,0,54}
\definecolor{bluex}{RGB}{0, 128, 128}

\usepackage[accsupp]{axessibility}  % Improves PDF readability for those with disabilities.

\usepackage[breaklinks,colorlinks,citecolor=eccvblue]{hyperref}
\usepackage{orcidlink}

\begin{document}

\title{LENS: Adaptive Spatio-Temporal Zooming for Keyframe Sampling in Long-Form Videos} 

\titlerunning{LENS: Adaptive Spatio-Temporal Zooming in Long-Form Videos}

\author{Ce Zhang \quad
Jinxi He \quad
Katia Sycara \quad
Yaqi Xie}

\authorrunning{C.~Zhang et al.}

\institute{Robotics Institute, Carnegie Mellon University\\ \texttt{\{cezhang, ginh, katia, yaqix\}@cs.cmu.edu}}

\maketitle
\input{sec/0_abstract}    
\input{sec/1_intro}
\input{sec/2_related}
\input{sec/3_method}
\input{sec/4_experiment}

\input{sec/5_conclusion}
% \clearpage

\bibliographystyle{splncs04}
\bibliography{main}

\input{sec/X_appendix}
\end{document}

%% file: sec/0_abstract.tex
\begin{abstract}

Despite rapid progress in Multi-modal Large Language Models (MLLMs), understanding long-form videos is still bottlenecked by limited context windows. While recent keyframe sampling methods attempt to mitigate this by distilling video inputs into a compact set of query-relevant frames, navigating the vast spatio-temporal search space remains challenging, as spatial detail and temporal coverage often conflict. To address this, we introduce LENS, a training-free keyframe sampling framework that dynamically decides when to zoom in for fine-grained details and when to zoom out for broader context based on the text query. Concretely, LENS adaptively allocates a limited frame budget between spatial zoom-ins, which highlight query-relevant regions within individual frames, and temporal zoom-outs, which expand the temporal scope through multi-frame aggregation, enabling the model to reason across multiple granularities while capturing both high-fidelity details and long-range context. Across diverse long-form video benchmarks, LENS consistently outperforms prior state-of-the-art keyframe sampling methods and delivers substantial gains over uniform sampling, improving Video-MME accuracy from 53.3\% to 60.7\% with Qwen2.5-VL.
% Extensive evaluations across diverse long-form video benchmarks show that LENS consistently surpasses prior state-of-the-art keyframe sampling approaches and delivers significant gains over uniform sampling, improving accuracy on the Video-MME benchmark from 53.3\% to 60.7\% on Qwen-2.5-VL. 
Code is available at \url{https://github.com/zhangce01/LENS}.
  % \keywords{Efficient Video Understanding \and Keyframe Sampling}
  \keywords{Long-form Video Understanding \and Keyframe Sampling
  \and Multi-modal Large Language Models \and Spatio-temporal Reasoning}
\end{abstract}

%% file: sec/1_intro.tex
\section{Introduction}
\label{sec:intro}

\begin{figure}[t]
\centering
\includegraphics[width=\linewidth]{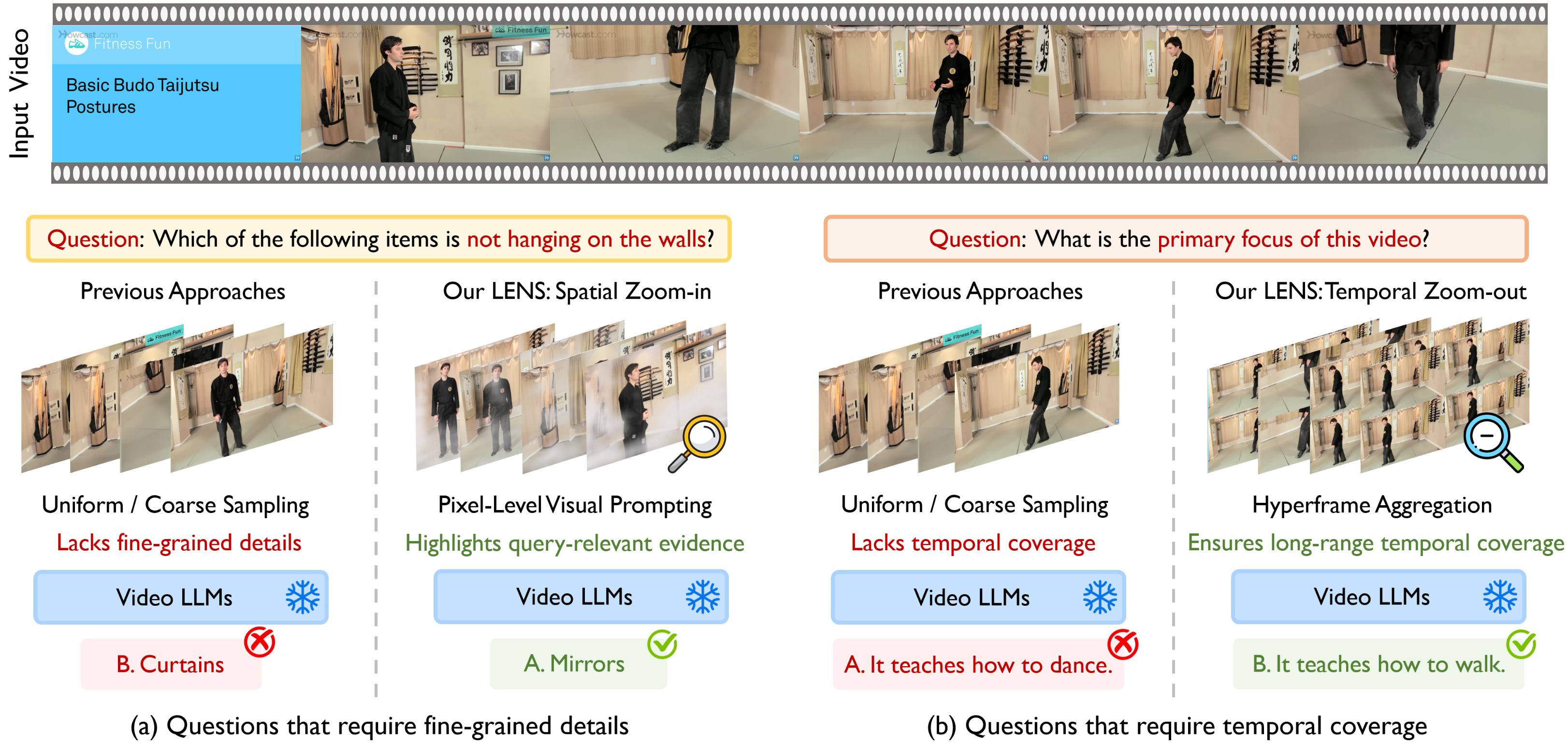}
\vspace{-15pt}
\caption{\textbf{Illustration of LENS for multi-granular video understanding}. For detail-oriented questions (\textit{Left}), conventional uniform/coarse sampling fails to capture fine-grained visual evidence, whereas our spatial zoom-in highlights query-relevant regions and enables correct prediction. For event-level understanding questions (\textit{Right}), prior approaches suffer from limited temporal coverage, while our temporal zoom-out expands long-range temporal context, allowing the model to identify the global temporal structure. Together, these results demonstrate that LENS adaptively allocates visual evidence across spatial and temporal dimensions to improve video reasoning.}
\label{fig:teaser}
\vspace{-15pt}
\end{figure}

% A straightforward strategy for handling long videos is uniform sampling, where frames are selected at regular intervals to satisfy context constraints. Although efficient, this approach can miss subtle yet important visual details, as it treats all temporal segments as equally informative.

% To address the context bottleneck, prior work has explored keyframe sampling strategies that aim to compress long videos into a compact set of representative frames. These approaches typically prioritize either frame-level relevance to the query or temporal diversity to preserve event coverage. However, selecting an optimal subset remains inherently challenging because long-form videos span a vast spatio-temporal search space, where capturing fine-grained visual evidence often comes at the expense of maintaining sufficient temporal context. As a result, fixed-granularity sampling strategies struggle to simultaneously preserve local details and global temporal structure. We argue that effective long-form video reasoning requires adaptively allocating computational resources across both spatial and temporal dimensions, allowing models to dynamically focus on either detailed visual cues or broader contextual signals depending on the query.

The emergence of Multi-modal Large Language Models (MLLMs) has expanded the scope of visual reasoning by bridging high-level linguistic understanding with complex visual perception~\cite{bai2025qwen25vl,zhang2026instruction,liu2023visual,zhang2026evolving,zhang2025self,zhang2026progresslm,zhang2024dual}. By integrating robust vision encoders with the reasoning capabilities of large language models, these systems demonstrate strong generalization across a wide range of tasks, from zero-shot captioning to multi-modal dialogue. Despite these advances, extending MLLMs to long-form video understanding remains a significant challenge~\cite{jin2024chat,lin2024video,liu2024end,zhang2024simple,zhang2026vscan,zhang2025vlm2}. Unlike images or short clips, long videos consist of thousands of frames with complex temporal dependencies, while current MLLMs can process only a limited number of visual tokens due to the limited context window. For example, a 30-minute video at 30 fps contains roughly 54,000 frames, which can translate to tens of millions of visual tokens when processed by models such as LLaVA-OneVision~\cite{li2025llavaonevision}, far exceeding typical context limits. As a result, directly feeding full videos into the model is computationally prohibitive, necessitating mechanisms that enable MLLMs to efficiently handle video understanding tasks.

To address the context bottleneck, a natural strategy is to select only a small subset of frames from the video as visual inputs for downstream reasoning. The efficacy of long-form video understanding is therefore intrinsically tied to the quality of this selection; any omission of pivotal moments leaves the model with insufficient information to address the query. Despite this, many state-of-the-art MLLMs~\cite{zhang2025llavavideo,li2025llavaonevision,zhang2026spacenum,pi2024mllm, bi2026think, bi2025reasoning} still rely on naive uniform sampling strategies that select frames at fixed intervals, which can miss task-critical content when relevant evidence is sparse or localized in time. This highlights the practical importance of \textit{keyframe sampling}, where selecting informative and query-relevant frames is essential for effective long-form video reasoning. However, determining which frames are truly informative is inherently challenging, as relevant visual cues may be sparse, subtle, or distributed across distant segments of the video.

% discuss prior work
To tackle this problem, recent works have explored various keyframe sampling strategies aimed at identifying frames that are most informative for answering a given query. For example, BOLT~\cite{liu2025bolt} and AKS~\cite{tang2025adaptive} use image–text matching models to estimate query–frame relevance scores, then apply inverse transform sampling or adaptive judge-and-split procedures to balance relevance with temporal diversity. 
In contrast, T$^*$~\cite{ye2025re} utilizes object detection models to identify inferred cue objects and iteratively refine its frame selection process. However, these approaches primarily operate at the frame level and treat frames as independent sampling units, focusing on selecting which frames to retain rather than how visual evidence should be allocated across spatial and temporal dimensions. Consequently, their fixed-granularity design limits flexibility under diverse query demands, as selected frames alone may not provide sufficient spatial precision or temporal coverage, as illustrated in Figure~\ref{fig:teaser}.

In this work, we introduce LENS, a novel framework for keyframe sampling that dynamically adjusts its focus between fine-grained spatial details and long-range temporal context. Our key insight is that different queries require different forms of visual evidence: object-centric questions rely on fine-grained local cues, whereas event-level reasoning depends on broader temporal context. Instead of selecting a static subset of frames, LENS dynamically allocates frame budgets to emphasize either detailed visual regions or wider temporal coverage according to the query. As illustrated in Figure~\ref{fig:teaser}, LENS enables multi-granular reasoning that more effectively aligns visual evidence with the demands of the task.

Specifically, LENS realizes this adaptive strategy through two complementary operations: spatial zoom-in and temporal zoom-out.\footnote{We use ``zoom'' as an analogy along two orthogonal axes rather than in the strict optical sense. \emph{Spatial zoom-in} narrows the spatial field of view within a frame to amplify local detail, whereas \emph{temporal zoom-out} widens the temporal field of view by aggregating neighboring frames.} The spatial zoom-in module highlights query-relevant regions within selected frames to capture fine-grained visual cues, while the temporal zoom-out module aggregates neighboring frames to expand the temporal field of view and preserve contextual continuity. A lightweight query-aware controller dynamically determines how to allocate the frame budget between these two branches, allowing the model to flexibly prioritize local precision or global context depending on the task. Notably, LENS is training-free and plug-and-play, making it directly applicable to both open-source and proprietary MLLMs without additional fine-tuning.

\looseness=-1
We comprehensively evaluate our approach on multiple long-form video question answering benchmarks and demonstrate that LENS consistently outperforms state-of-the-art training-free sampling methods across frame budgets and backbones. Qualitative analyses further highlight LENS’s ability to flexibly balance spatial precision and temporal coverage, showing that it dynamically allocates frame budgets between spatial zoom-ins and temporal zoom-outs based on the text query, prioritizing fine-grained visual regions for detail-oriented questions while expanding temporal context for queries that require broader event-level understanding.

% We evaluate LENS on multiple long-form video benchmarks and demonstrate consistent improvements over state-of-the-art keyframe sampling approaches. Notably, when applied to Qwen-2.5-VL, LENS improves accuracy on the Video-MME benchmark from 53.3% to 60.7%, while also outperforming uniform sampling by a significant margin. These results indicate that adaptive spatio-temporal allocation is a simple yet effective strategy for improving long-video reasoning.

Our contributions are summarized as follows:
\setlist[itemize]{itemsep=2pt, topsep=2pt}
\begin{itemize}
    \item[$\bullet$] We propose LENS, a novel training-free framework that adaptively partitions a limited frame budget between spatial zoom-in and temporal zoom-out operations, enabling multi-granular reasoning that balances fine-grained visual details with long-range temporal context.
    \item[$\bullet$] \looseness=-1 We devise spatial zoom-in and temporal zoom-out mechanisms that refine local evidence through query-conditioned visual prompting and capture global video structure via graph-based reasoning with hyperframe aggregation.
    \item[$\bullet$] We validate the effectiveness of LENS through comprehensive experiments on three long-form video understanding benchmarks, demonstrating that it consistently outperforms state-of-the-art keyframe sampling approaches.
\end{itemize}

%% file: sec/2_related.tex
\section{Related Work}
\label{sec:related}

\looseness=-1
\textbf{Long-Form Video Understanding}. Recent progress in multi-modal learning has enabled large language models to perceive videos by augmenting them with visual encoders, converting frames into visual tokens that can be processed alongside text for a wide range of video understanding tasks~\cite{bai2025qwen25vl,cheng2024videollama,fu2025video,zhu2024minigpt,tang2025video,bi2026think,bi2025verify}. However, early approaches struggle to handle hour-long videos in a single pass, as the resulting token sequences quickly exceed the available context length~\cite{xu2025slowfastllava,zhang2025long,jin2024chat,cheng2024videollama,ren2024timechat}.
To address this challenge, recent work has focused on the fundamental problem of scaling long-context MLLMs and improving their temporal reasoning capabilities~\cite{santos2025video,xu2025slowfastllava}. For instance, LongVU~\cite{shen2025longvu} introduces spatiotemporal adaptive compression to reduce temporal/spatial redundancy and fit many frames into a fixed window, and LongVILA~\cite{chen2025longvila} provides a full-stack solution that extends context length and performs long-video supervised fine-tuning with system-level optimizations. Another line of work tackles long-form video understanding from an inference-time perspective by designing LLM-based agents equipped with external tools to iteratively explore videos and gather relevant evidence. Representative methods include VideoAgent~\cite{wang2024videoagent}, VideoLucy~\cite{zuo2025videolucy}, and VideoTree~\cite{wang2025videotree}, which dynamically select tools (\eg, frame captioning, temporal navigation) to decompose complex queries into sequential reasoning steps. More recently, retrieval-augmented generation (RAG) techniques have been adopted to enhance LLMs by building external knowledge bases and retrieving relevant information to support long-context reasoning~\cite{luo2025videorag,xue2025adavideorag,shen2025vgent}. 
However, the heavy computational overhead introduced by iterative tool calling hinders their scalability and practical deployment.

\vspace{5pt}
\looseness=-1
\noindent\textbf{Video Keyframe Sampling}. Orthogonal to these lines of work, we focus on the problem of \textit{video keyframe sampling}, which aims to select a compact, query-relevant subset of frames that maximizes useful evidence~\cite{yu2023self,yan2023unloc,hu2025m,buch2025flexible,bi2025diagnosing,tang2026video}. Early methods such as BOLT~\cite{liu2025bolt} and AKS~\cite{tang2025adaptive} frame keyframe selection as a trade-off between relevance and temporal diversity, leveraging inverse transform sampling or adaptive judge-and-split procedures to balance prompt–frame relevance with coverage. T$^*$~\cite{ye2025re} advances this line of work by introducing a temporal search mechanism that detects visual cues and progressively refines sampling, upsampling relevant temporal or spatial regions using object detection models. Further, Q-Frame~\cite{zhang2025q} performs query-aware frame selection with multi-resolution adaptation, allocating higher resolution to the most informative frames while keeping the overall token budget fixed. However, these approaches operate at a single granularity, focusing on which frames to select rather than how evidence should be distributed across spatial and temporal scales. 
In contrast, we introduce spatial zoom-in and temporal zoom-out to enable multi-granular reasoning that adaptively balances spatial precision and temporal coverage according to the query.

%% file: sec/3_method.tex
\section{Method}
\label{sec:method}

\begin{figure}[t]
\centering
\includegraphics[width=\linewidth]{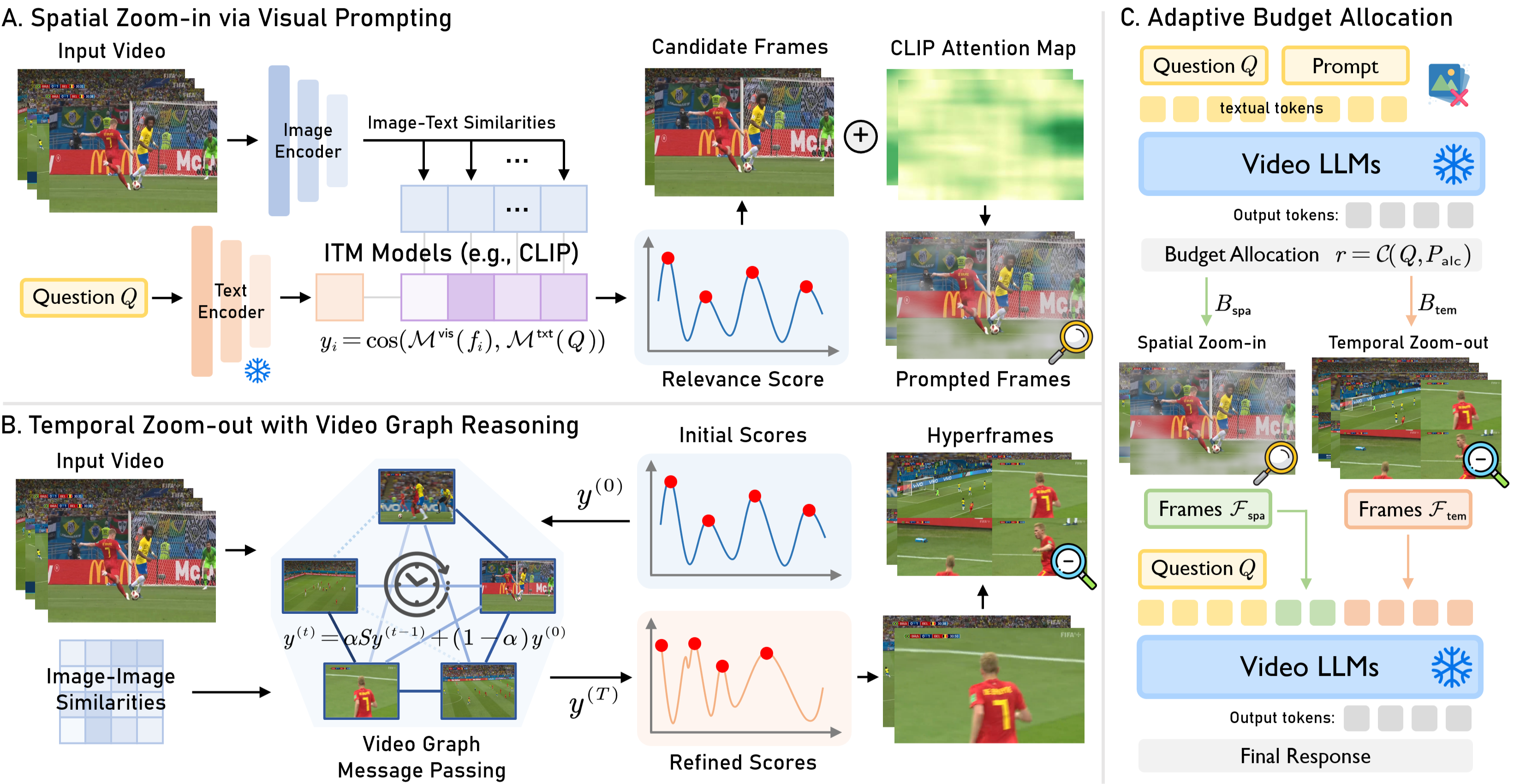}
\vspace{-15pt}
\caption{\textbf{Overview of LENS}. (A) Spatial zoom-in selects query-relevant frames using image–text similarity and enhances fine-grained evidence via CLIP attention–based visual prompting.
(B) Temporal zoom-out constructs a video graph to model inter-frame relationships and propagates relevance scores through message passing, enabling long-range reasoning and the selection of context-rich hyperframes.
(C) An adaptive budget allocator dynamically distributes frame budgets between spatial and temporal branches, whose outputs are combined for final reasoning with a video LLM.}
\label{fig:overview}
\vspace{-10pt}
\end{figure}

In this work, we introduce LENS (Figure~\ref{fig:overview}), a novel training-free framework that adaptively samples frames by performing spatial zoom-ins and temporal zoom-outs, enabling the model to reason across multi-granular scales to achieve both high-fidelity detail and comprehensive temporal coverage.

\subsection{Preliminaries}
\noindent\textbf{MLLM-Based Video Understanding}.
Consider a video question answering task with an input video $V$ consisting of $T$ discrete frames $\{f_1, f_2, \cdots, f_T\}$ and an associated natural language query $Q$. To generate a linguistically coherent and visually grounded response, the MLLM employs a visual encoder $\mathcal{E}$ to extract latent features from each frame and a feature projector $\mathcal{P}$ to map these features into a sequence of visual tokens $\mathbf{x}_V = \mathcal{P}\left(\mathcal{E}(f_1), \mathcal{E}(f_2), \cdots, \mathcal{E}(f_T)\right)$. These visual tokens are concatenated with the tokenized textual query $\mathbf{x}_T$ and fed into an LLM decoder for auto-regressive next-token generation. Formally, at each decoding step $j$, the next token $y_j$ is sampled from the output probability distribution $p_{\theta}$ conditioned on the multimodal context and previously generated tokens $\mathbf{y}_{<j}$, represented as $y_j \sim p_{\theta}(y_j \mid \mathbf{x}_V, \mathbf{x}_T, \mathbf{y}_{<j})$.

\vspace{5pt}
\noindent\textbf{Keyframe Sampling}. Processing the entire sequence of $T$ frames is often computationally prohibitive due to the quadratic complexity of self-attention in the LLM decoder. To mitigate this, a sampling strategy $\mathcal{S}$ is typically employed to select a subset of $k$ keyframes $\{f_{s_1}, f_{s_2}, \cdots, f_{s_k}\}$, where $k \ll T$. By reducing the temporal redundancy of the input video, keyframe sampling ensures that the visual tokens $\mathbf{x}_V$ remain within the LLM's effective context window while preserving the essential semantic content required for accurate video understanding.

\subsection{Spatial Zoom-in via Visual Prompting}
\looseness=-1
In long-form videos, query-relevant evidence is frequently sparse and localized, yet most approaches~\cite{liu2025bolt,tang2025adaptive} rely on coarse frame-level representations that overlook such spatial nuances. We introduce a \textit{pixel-level visual prompting} module that enhances spatial granularity by applying a query-conditioned attention mask to each frame. Instead of treating visual inputs uniformly, the module emphasizes semantically relevant and visually salient regions, effectively performing a \textit{spatial zoom-in} that preserves visual context while capturing fine-grained cues.

\vspace{5pt}
\looseness=-1
\noindent\textbf{Candidate Frame Selection}. To identify a subset of query-relevant frames $\mathcal{F}_{\mathsf{spa}}'$ for subsequent visual prompting, we first extract a pool of candidate frames from the input video. We utilize a pre-trained image–text alignment model $\mathcal{M}$ (\eg, CLIP~\cite{radford2021learning} or BLIP~\cite{li2022blip}) to measure the relevance between each frame $f_i$ and the query $Q$ via cosine similarity, defined as
\begin{equation}
\label{eq:cosine}
y_i = \text{cos}(\mathcal{M}^{\mathsf{vis}}(f_i), \mathcal{M}^{\mathsf{txt}}(Q)), \quad i \in\{1, \cdots, T\},
\end{equation}
where $\mathcal{M}^{\mathsf{vis}}$ and $\mathcal{M}^{\mathsf{txt}}$ denote the visual and textual encoders, respectively. To prevent selected frames from clustering within a narrow temporal window, we implement a watershed algorithm~\cite{haralick1987image} on the similarity curve to partition the video into distinct temporal segments. We then select the single highest-scoring frame from each segment as a representative. This ensures the candidate set $\mathcal{F}_{\mathsf{spa}}'$ spans the full video timeline rather than concentrating around a single event.

\vspace{5pt}
\noindent\textbf{Query-Guided Visual Prompting}. To highlight query-relevant regions within each frame of the candidate set $\mathcal{F}_{\mathsf{spa}}'$, we leverage cross-modal attention signals within a ViT-based CLIP~\cite{radford2021learning} model $\mathcal{M}_{\mathsf{CLIP}}$ to derive spatial attribution maps that quantify the semantic relevance between the textual query and individual image patches, which enable us to localize fine-grained evidence while preserving the global visual context.

To decompose the global image–text similarity of each frame $f$ into patch-level relevance scores, we formalize the final $\texttt{[CLS]}$ representation as a linear aggregation of the value flows through the preceding attention and MLP layers following prior works \cite{gandelsman2024interpreting}:
\begin{equation}
\mathcal{M}_{\mathsf{CLIP}}^{\mathsf{vis}}(f)\!=\!\phi\left([\mathbf{z}^0]_{\mathsf{cls}}\right)\!+\!\sum_{l=1}^L \phi\left(\left[\mathsf{MSA}^l (\mathbf{z}^{l-1})\right]_\mathsf{cls}\right) + \sum_{l=1}^L\phi\left(\left[\mathsf{MLP}^l (\mathbf{\tilde{z}}^{l})\right]_\mathsf{cls}\right)
\end{equation}
where $L$ denotes the total number of transformer layers, $\mathbf{z}^{l-1}$ and $\tilde{\mathbf{z}}^{l}$ represent the token sequences prior to the $l$-th attention and MLP blocks respectively, and where $\phi(\cdot)$ denotes the final linear projection (with normalization) that maps visual features into the joint embedding space.
Prior work~\cite{liu2023visual,yu2024attention} shows that the final similarity is primarily governed by the deepest Multi-Head Self-Attention (MSA) layers, i.e., $l \in \{L',\dots,L\}$, while the initial embedding $\mathbf{z}^0$ and MLP outputs act mainly as residual contributions.

\looseness=-1
We further decompose the outputs of the MSA blocks as a weighted summation over $H$ attention heads and $N$ input tokens:
\begin{equation}
    \sum_{l=L'}^L \phi\left(\left[\mathsf{MSA}^l (\mathbf{z}^{l-1})\right]_\mathsf{cls}\right) = \sum_{l=L'}^L\sum_{h=1}^H \sum_{n=0}^N \phi\left(\alpha_n^{(l,h)} W^{(l,h)} z_n^{l-1}\right) = \sum_{l=L'}^L\sum_{h=1}^H \sum_{n=0}^N s^{(n, l, h)},
\end{equation}
where $\alpha_n^{(l,h)}$ represents the attention weight assigned to the $n$-th token by the $h$-th head at layer $l$ relative to the $\texttt{[CLS]}$ token. The term $\mathbf{V}_n^{(l,h)} = W_v^{(l,h)} z_n^{l-1}$ denotes the linear value projection of the $n$-th token. 
The term $s^{(n, l, h)}$ encapsulates the individual contribution of the $n$-th token to the final representation through a specific head and layer. By aggregating these contributions across all terminal layers and attention heads, we derive the final relevance score for each image patch $n \in \{1, \dots, N\}$ as $\mathbf{c}_n = \sum_{l=L'}^L \sum_{h=1}^H s^{(n, l, h)}$. We then compute a query-conditioned relevance score by measuring alignment with the text embedding $r_n = \langle \mathbf{c}_n, \mathcal{M}_{\mathsf{CLIP}}^{\mathsf{txt}}(Q) \rangle$, where $\langle \cdot, \cdot \rangle$ denotes the inner product. Intuitively, this score measures how strongly each patch contributes in the direction of the query embedding.

\vspace{5pt}
\noindent\textbf{Saliency-Based Visual Prompting}. Empirically, these query-conditioned relevance scores are most effective when the query $Q$ specifies concrete objects or actions, as they can precisely localize regions aligned with explicit semantics. However, their discriminative capability weakens for more general queries, where relevance is not anchored to particular entities and instead depends on broader contextual cues. To mitigate this limitation, we introduce a complementary visual-only attribution that leverages the attention received from the $\texttt{[CLS]}$ token in the final transformer layer, which has been shown to encode global contextual information~\cite{yang2025visionzip,zhang2026vscan}.
Specifically, let $\mathbf{A}^{(L,h)} \in \mathbb{R}^{(N+1)\times(N+1)}$ denote the attention matrix of the $h$-th head in the final layer $L$.
We compute the visual-only contribution score for a specific token $n$ by aggregating the attention weights from $\texttt{[CLS]}$ across heads:
\begin{equation}\hat{r}_n = \frac{1}{H} \sum_{h=1}^H \mathbf{A}_{\mathsf{cls}\rightarrow n}^{(L,h)}, \quad n \in \{1, \dots, N\}.\end{equation}
To ensure the final mask captures both explicit semantic targets and general foreground saliency cues, we combine the query-conditioned relevance $r_n$ and the query-agnostic importance $\hat{r}_n$ using a soft OR fusion: $r_n^{\text{final}} = 1 - (1 - r_n)(1 - \hat{r}_n)$,
which increases whenever either signal indicates high relevance.

The fused token scores are arranged into an attribution map on the token grid and resized to the original image resolution to obtain a raw heatmap. To reduce discretization artifacts from patch-level tokens, we apply a mean filter to produce smoother spatial transitions that better align with semantic boundaries. 
This refined heatmap is then overlaid onto the original frames as an alpha mask, yielding a set of visually prompted frames $\mathcal{F}_{\mathsf{spa}}$, which effectively suppress irrelevant background noise and amplify query-relevant foreground regions.

\subsection{Temporal Zoom-out with Video Graph Reasoning}
Complementarily, long-form reasoning also requires understanding broader temporal context beyond isolated frames. To this end, we introduce a parallel branch with a temporal zoom-out mechanism that expands the temporal receptive field by aggregating information from a wider frame neighborhood. Instead of focusing on local evidence, this branch summarizes global temporal dynamics and contextual cues, enabling the model to capture long-range dependencies. By integrating fine-grained spatial evidence with coarse-grained temporal context, the framework achieves a more holistic understanding of video content.

\vspace{5pt}
\noindent\textbf{Video Graph Reasoning}. To explicitly model the temporal structure and semantic relationships among frames, we represent the input video as an undirected graph $\mathcal{G}=(\mathcal{V}, \mathcal{E})$, where each node corresponds to a frame representation.  We compute pairwise distances between frames using visual similarity measures such as CLIP image–image similarity~\cite{radford2021learning} or SSIM~\cite{wang2004image}. Specifically, the distance between two frames is defined as $d(f_i, f_j) = \left(1-\textsf{Sim}(f_i, f_j)\right) / 2\in[0,1]$, where $\textsf{Sim}(\cdot,\cdot)$ denotes the similarity score. We then sort all pairwise distances in ascending order and iteratively connect frame pairs following this order until the resulting graph becomes fully connected. Intuitively, this yields a sparse yet globally coherent graph that captures intrinsic video structure.

\looseness=-1
For all connected edges, we define the affinity matrix $W$ using a Gaussian kernel to represent the pairwise similarity between frames:
\begin{equation}
W_{ij} =
\begin{cases}
\exp\left( -\frac{d^2(f_i, f_j)}{2\sigma^2} \right) & \text{if } i \neq j, \\
0 & \text{if } i = j,
\end{cases}
\end{equation}
where $W_{ii}=0$ effectively prevents self-loops. We then compute the symmetrically normalized adjacency matrix $S = D^{-1/2} W D^{-1/2}$, where $D$ is the diagonal degree matrix with $D_{ii} = \sum_j W_{ij}$. Using this normalized structure, we perform iterative message passing until convergence according to the following update rule:
\begin{equation}
\label{kkk}
    y^{(t)} = \alpha S y^{(t-1)} + (1-\alpha)y^{(0)},
\end{equation}
where $y^{(0)}=y$ represents the initial image-text similarity scores between each frame and the text query $Q$ in Equation~(\ref{eq:cosine}), and $\alpha \in (0, 1)$ is a balancing hyperparameter that controls the trade-off between the local unary priors and the global graph structure. While Equation~(\ref{kkk}) admits a closed-form solution with $\mathcal{O}(T^3)$ complexity ($T$: total frames), we use the iterative form with $\mathcal{O}(K|\mathcal{E}|)$ cost ($|\mathcal{E}|$: edges, $K$: iterations), which is empirically negligible compared to MLLM processing time.
Formally, since the symmetrically normalized matrix $S$ has eigenvalues bounded within $[-1, 1]$, the message-passing sequence $\{y^{(t)}\}$ is a convergent Neumann series, which reaches a unique steady state. Concretely, $\{y^{(t)}\}$ converges to $y^*=(1-\alpha)(I-\alpha S)^{-1}y^{(0)}$ (see Zhou \etal\cite{zhou2003learning} for a rigorous proof). 

Intuitively, the refined scores obtained after the iterative updates produce a refined similarity landscape where the relevance of each frame is reinforced by the underlying video structure. Finally, we similarly apply watershed sampling over the refined scores to select another set of candidate frames $\mathcal{F}'_{\mathsf{tem}}$ that are globally representative of the video's semantics.

\vspace{5pt}
\noindent\textbf{Hyperframe Aggregation}. Given the selected candidate frames $\mathcal{F}'_{\mathsf{tem}}$, we construct \emph{hyperframes} to enrich temporal context while keeping the token budget fixed. Specifically, for each candidate frame, we concatenate its neighboring $\tau -1$ frames along the spatial axis to form a composite frame. The resulting hyperframe is then downsampled to a lower spatial resolution, trading spatial detail for broader temporal coverage, yielding the hyperframe set $\mathcal{F}_{\mathsf{tem}}$. This design enables the model to capture short-range motion and contextual cues around key moments without introducing additional frames, thereby improving temporal reasoning while maintaining computational efficiency.

\subsection{Adaptive Keyframe Budget Allocation}
The spatial and temporal branches described above provide complementary evidence: spatial zoom-in captures fine-grained local cues, while temporal zoom-out summarizes long-range context via video graph reasoning and hyperframes. However, different queries demand different levels of detail and contextual breadth. To reconcile this variability, we introduce a query-aware allocation mechanism that dynamically distributes a fixed frame budget $B$ between the two branches.

Given a natural language query $Q$ and an allocation prompt $P_\mathsf{alc}$, we employ the MLLM as a lightweight controller $\mathcal{C}$ that predicts an allocation ratio\footnote{See the Appendix for the prompt details.}
\begin{equation}
r = \mathcal{C}(Q, P_\mathsf{alc}), \qquad 
B_{\mathsf{spa}} = \Big\lfloor rB + \tfrac{1}{2} \Big\rfloor, \qquad 
B_{\mathsf{tem}} = B - B_{\mathsf{spa}},
\end{equation}

where $r \in [0,1]$ denotes the proportion of frames assigned to spatial zoom-in, and $B_{\mathsf{spa}}$ and $B_{\mathsf{tem}}$ specify the frame budgets for the spatial and temporal branches, respectively, i.e., $|\mathcal{F}_{\mathsf{spa}}| = B_{\mathsf{spa}}$ and $|\mathcal{F}_{\mathsf{tem}}| = B_{\mathsf{tem}}$. The final frame set is then formed by merging the two branches $\mathcal{F} = \mathsf{Sort}_{t}\!\left(\mathcal{F}_{\mathsf{spa}} \cup \mathcal{F}_{\mathsf{tem}}\right)$, where $\mathsf{Sort}_{t}(\cdot)$ orders frames according to their original temporal indices to preserve chronological consistency. These frames are then provided to the MLLM as visual inputs for multi-modal reasoning.

%% file: sec/4_experiment.tex
\section{Experiments}
\label{sec:experiment}

\input{table/main_table}

In this section, we validate the effectiveness of our keyframe sampling approach on various standard long video understanding benchmarks.

\subsection{Experimental Settings}
\looseness=-1
\textbf{Models and Benchmarks}.To evaluate our approach, we sample keyframes and process them using a suite of state-of-the-art MLLMs. Specifically, we employ Qwen2.5-VL \cite{bai2025qwen25vl}, LLaVA-OneVision \cite{li2025llavaonevision}, and GPT-5-Mini as our foundational models for video understanding. We conduct extensive experiments on three long-form video understanding benchmarks: 
\setlist[itemize]{itemsep=2pt, topsep=2pt}
\begin{itemize}
    \item[$\bullet$] \textbf{Video-MME~\cite{fu2025video}} is a widely used benchmark designed to evaluate MLLMs across various video lengths, requiring the integration of both short-term temporal cues and long-term semantic context;
    \item[$\bullet$] \textbf{LongVideoBench (LVB)~\cite{wu2024longvideobench}} evaluates a model's ability to perform complex reasoning and information retrieval over extended durations;
    \item[$\bullet$] \textbf{MLVU~\cite{zhou2025mlvu}} covers a diverse range of disciplines and tasks, providing a holistic assessment of a model’s capacity to process long-sequence video data in specialized domains.
\end{itemize}
Collectively, these benchmarks feature videos ranging from several seconds to hours, providing a comprehensive evaluation of the effectiveness of our approach in selecting the most representative keyframes for long-form video understanding.

% (1) \textbf{Video-MME~\cite{fu2025video}} is a widely used benchmark designed to evaluate MLLMs across various video lengths, requiring the integration of both short-term temporal cues and long-term semantic context; (2) \textbf{LongVideoBench (LVB)~\cite{wu2024longvideobench}} evaluates a model's ability to perform complex reasoning and information retrieval over extended durations; (3) \textbf{MLVU~\cite{zhou2025mlvu}} covers a diverse range of disciplines and tasks, providing a holistic assessment of a model’s capacity to process long-sequence video data in specialized domains. 

\vspace{5pt}
\noindent\textbf{Baselines}. As a simple baseline, we include results from a uniform sampling baseline, where frames are selected at equal time intervals. Furthermore, we compare our performance against three state-of-the-art keyframe selection methods: BOLT~\cite{liu2025bolt}, AKS~\cite{tang2025adaptive}, and Q-Frame~\cite{zhang2025q}. To ensure a fair comparison, we report the performance of these baselines based on our re-implementation.

\vspace{5pt}
\noindent\textbf{Implementation Details}. For all evaluations, we adhere to the default query formats and prompts specified by the respective benchmarks to ensure consistency. Following the evaluation protocol in prior work~\cite{tang2025adaptive,zhang2025q}, we configure our method to sample 8, 16, and 32 frames from the input video. For the temporal zoom-out module, we set $\sigma = 0.3$, $\alpha = 0.8$, and $\tau = 4$ as the default hyperparameters for hyperframe generation; these choices are supported by ablation studies reported in the Appendix. To prevent redundancy between the two branches, we explicitly constrain the temporal zoom-out module to avoid selecting frames already chosen by spatial zoom-in. All experiments are conducted on a server equipped with 8$\times$ 48GB NVIDIA RTX 6000 Ada GPUs.

\subsection{Results and Discussions}
\noindent\textbf{Results on Qwen-2.5-VL}. Table~\ref{tab:main-results} presents a comprehensive comparison between LENS and existing training-free keyframe sampling strategies under varying frame budgets. Across all budgets (8, 16, and 32 frames), LENS consistently achieves the best performance, demonstrating the effectiveness of adaptive multi-granular keyframe selection. Under the strict 8-frame constraint, LENS attains an overall accuracy of 60.7\% on Video-MME, surpassing the second-best baseline Q-Frame by 2.0\%. The gains are consistent across short, medium, and long videos, indicating that even with limited visual evidence, the combination of spatial zoom-in and temporal zoom-out provides more informative frame selection than relevance-only or heuristic-based sampling.
As the frame budget increases, the advantage of LENS becomes more pronounced. At 16 frames, LENS reaches 63.1\% overall accuracy, improving over Q-Frame by 2.4\% and showing notable gains on medium and long videos, where modeling long-range dependencies becomes increasingly important. With 32 frames, LENS achieves 67.1\%, outperforming all baselines by a substantial margin. Overall, these results highlight that LENS is both budget-efficient and scalable, delivering consistent gains with limited frames while further amplifying improvements as more budgets becomes available.

LENS also demonstrates consistent improvements on LongVideoBench and MLVU. Under the 8-frame setting, LENS achieves 59.3\% on LVB and 63.7\% on MLVU, outperforming the strongest baseline Q-Frame by 2.7\% and 5.5\%, respectively. The gains remain stable as the frame budget increases, reaching 60.6\% on LVB and 66.7\ on MLVU with 32 frames. These results further demonstrate that LENS consistently selects more informative frames and generalizes well across different long-form video benchmarks.

\vspace{5pt}
\noindent\textbf{Results on Other MLLMs}. 
To verify that LENS generalizes beyond Qwen2.5-VL, we apply it to two additional backbones under the 8-frame budget. On LLaVA-OneVision, LENS improves the overall accuracy on Video-MME from 53.2\% to 58.7\%, with consistent gains on short, medium, and long videos. Similar trends are observed on other benchmarks, where LENS increases performance on LongVideoBench from 51.8\% to 57.5\% and on MLVU from 59.2\% to 66.3\%. On the GPT-5-Mini backbone, LENS also yields clear improvements, raising Video-MME overall accuracy from 64.8\% to 69.0\% and notably enhancing long-video performance (58.9\% vs.\ 65.6\%). Meanwhile, LENS improves LongVideoBench from 55.0\% to 61.7\% and MLVU from 49.1\% to 53.8\%. These results suggest that LENS serves as a model-agnostic, training-free plug-in that consistently enhances long-form video reasoning by providing complementary spatial evidence and temporal context under tight frame budgets.

\begin{figure}[t]
\centering
\includegraphics[width=\linewidth]{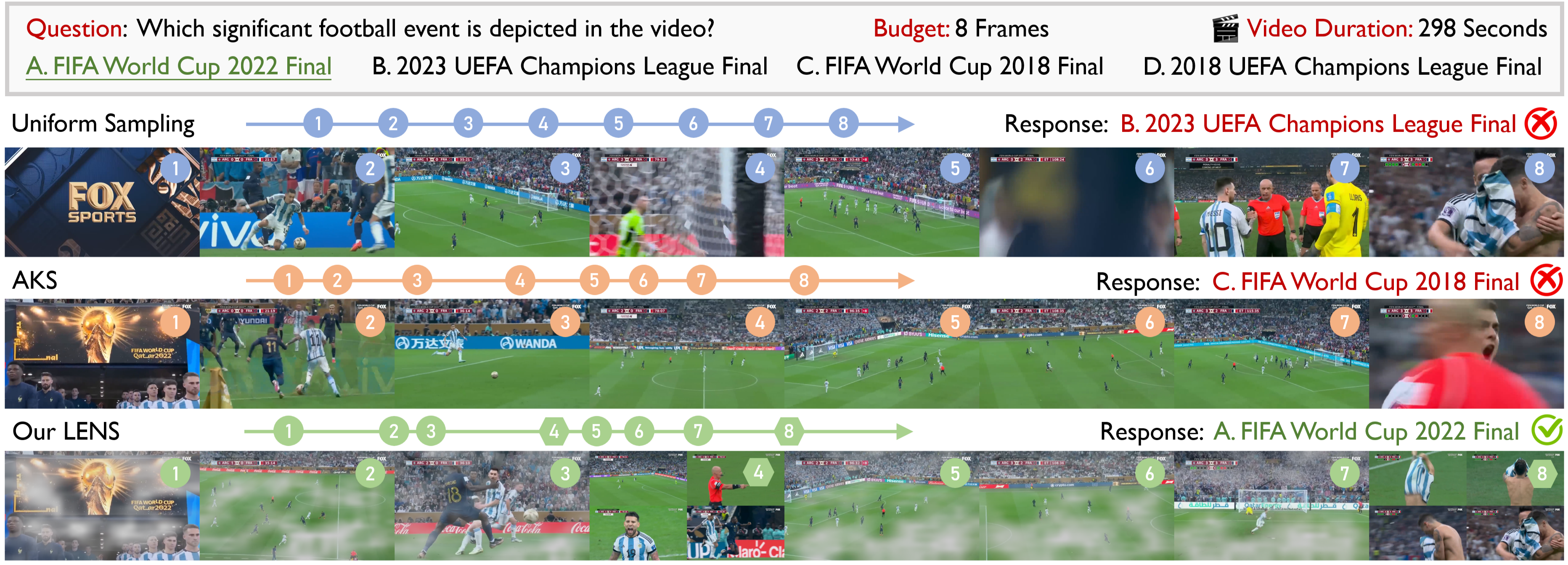}
\includegraphics[width=\linewidth]{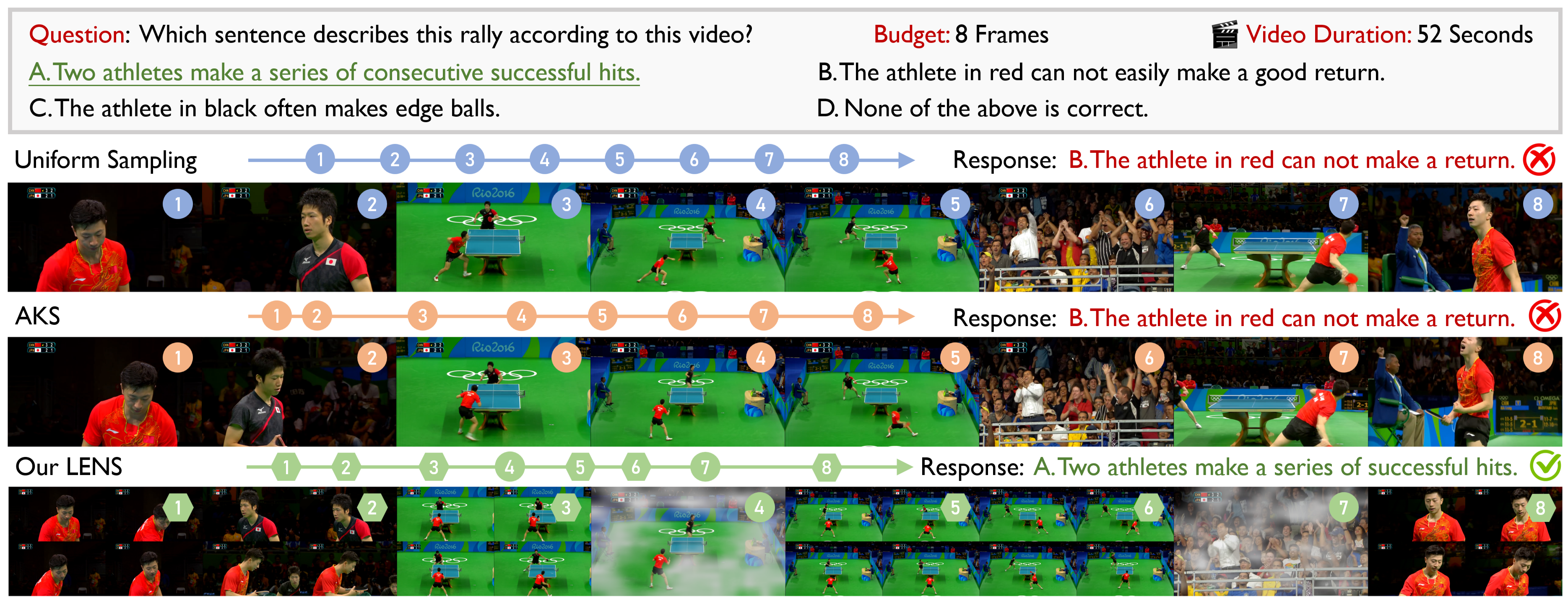}
\vspace{-18pt}
\caption{\textbf{Qualitative comparison on two representative samples from the Video-MME benchmark}~\cite{fu2025video}. The selected frames from LENS are compared against those obtained via uniform sampling and AKS~\cite{tang2025adaptive}. Note that frames obtained via spatial zoom-in are denoted by green circles, while those from temporal zoom-out are represented by green hexagons.}
\label{fig:qualitative}
\vspace{-15pt}
\end{figure}

\vspace{5pt}
\noindent\textbf{Qualitative Results}. Figure~\ref{fig:qualitative} presents two representative comparisons of frame selection on Video-MME. In the first (football event recognition), uniform sampling distributes frames evenly across the timeline but misses key discriminative cues, while AKS retrieves partially relevant moments yet fails to capture the decisive textual evidence at sufficient spatial resolution; consequently, both methods produce incorrect predictions. In contrast, LENS uses spatial zoom-in to emphasize fine-grained visual cues (e.g., the event-title region) and temporal zoom-out to preserve the overall match context, enabling the model to correctly identify the event as the \textit{FIFA World Cup 2022 Final}. The second comparison (table tennis rally understanding) further underscores the importance of multi-granular reasoning. Uniform sampling and AKS attend to isolated moments of the rally and overlook the sustained interaction between players, resulting in incorrect interpretations of the play. By adaptively allocating frames, LENS captures both localized player actions and the temporal continuity of the rally, allowing the model to correctly infer that the sequence depicts \textit{a series of consecutive successful hits}. Overall, these results show that LENS selects frames that are not only semantically relevant but also complementary across spatial and temporal scales, yielding more reliable reasoning than prior keyframe sampling strategies.

% \noindent\textbf{Qualitative Results}. Figure~\ref{fig:qualitative} presents two representative comparisons of frame selection on Video-MME. In the first example (football event recognition), uniform sampling distributes frames evenly across the timeline but misses key discriminative cues, while AKS retrieves partially relevant moments yet fails to capture the decisive textual evidence at a sufficient spatial resolution. Consequently, both methods lead to incorrect predictions. In contrast, LENS leverages spatial zoom-in to emphasize fine-grained visual cues (e.g., the event title region) and temporal zoom-out to preserve the overall match context, enabling the model to correctly identify the event as the \textit{FIFA World Cup 2022 Final}.

% The second example (table tennis rally understanding) further highlights the importance of multi-granular reasoning. Uniform sampling and AKS focus on isolated moments of the rally and overlook the sustained interaction pattern between players, resulting in incorrect interpretations of the play. By adaptively allocating frames, LENS captures both localized player actions and the temporal continuity of the rally, allowing the model to correctly infer that the sequence depicts \textit{a series of consecutive successful hits}.

% Overall, these qualitative results illustrate that LENS consistently selects frames that are not only semantically relevant but also complementary across spatial and temporal scales, leading to more reliable reasoning than prior keyframe sampling strategies.

\noindent\textbf{Additional Results}. Please refer to Appendix~\ref{appendix:exp} for more experimental results on EgoSchema~\cite{mangalam2023egoschema} and NExT-QA~\cite{xiao2021next}, which consistently demonstrates that our LENS can generalize to diverse video domains and reasoning types.

\subsection{More Discussions}

\noindent\textbf{Efficiency Analysis}. Table~\ref{table:efficiency} reports the total runtime and accuracy of different methods under an 8-frame budget, evaluated on all 2,700 Video-MME samples using eight NVIDIA RTX 6000 Ada GPUs. While LENS incurs a modest runtime increase over uniform sampling (11 h 50 min \vs 8 h 13 min), it achieves the highest accuracy of 60.7\%, corresponding to a substantial gain of +7.4\%. Compared to prior adaptive sampling approaches such as AKS and Q-Frame, LENS yields consistently larger improvements with only moderate additional computational overhead. This overhead primarily arises from the lightweight allocation controller, as well as the extra computation introduced by attention-based prompting in spatial zoom-in and graph-based reasoning in temporal zoom-out; however, these components remain computationally inexpensive in practice. Overall, these results confirm that LENS provides a strong balance between accuracy and efficiency, making it well-suited for real-world long-video applications.

\begin{wraptable}[11]{r}{0.55\textwidth}
\small
\begin{center}
\vspace{-40pt}
\caption{\textbf{Efficiency comparison on Video-MME~\cite{fu2025video}}. We report the total runtime, the achieved accuracy, and the performance gains compared to the uniform sampling baselines.}
\label{table:efficiency}
\vspace{-5pt}
\resizebox{\linewidth}{!}{
\begin{tabular}{lccc}
\toprule
Method  & Runtime & Accuracy & Gain \\ 
\midrule
Qwen-2.5-VL~\cite{bai2025qwen25vl}  &  8 h 13 min &  53.3&  -\\
AKS~\cite{tang2025adaptive} &9 h 55 min &  54.1&  +0.8 \\
Q-Frame~\cite{zhang2025q} &10 h 30 min &  58.7&  +5.4 \\
\rowcolor{gray!20}
\textbf{LENS (Ours)} &   11 h 50 min  & \textbf{60.7} & \textbf{+7.4} \\
\bottomrule
\end{tabular}
}
\end{center}
\end{wraptable}
% Notably, LENS remains substantially more efficient than recent LLM-agent and retrieval-augmented video reasoning approaches. For example, VideoRAG~\cite{luo2025videorag} and Vgent~\cite{shen2025vgent} require over 50 hours to process the full benchmark, highlighting the practicality of LENS as a lightweight and scalable solution for long-form video understanding.

Notably, LENS remains far more efficient than even recent LLM-agent and retrieval-augmented video reasoning methods. For instance, VideoRAG~\cite{luo2025videorag} and Vgent~\cite{shen2025vgent} each require over 50 hours on the same benchmark—more than 4$\times$ the runtime of LENS—underscoring its practicality as a lightweight solution for long-form video understanding.

\looseness=-1
We also provide the average per-query runtime breakdown on Video-MME, where numbers in parentheses denote the cumulative performance after adding each component. LENS takes 15.8s in total: 11.0s for standard MLLM processing (53.3), 0.8s for CLIP/BLIP similarity computation (56.6), 1.4s for visual prompting (59.6), 0.5s for graph reasoning (60.2), and 2.1s for budget allocation (60.7).

\input{table/main_ablation}

% \vspace{5pt}
\looseness=-1
\noindent\textbf{Ablation Studies}. Table \ref{tab:main2} evaluates the contribution of each component in LENS under a strict 8-frame budget. Starting from the vanilla Qwen2.5-VL baseline, incorporating query-aware frame scoring already yields notable improvements, where CLIP-based selection increases overall accuracy from 53.3\% to 56.6\%, and BLIP further improves it to 57.0\%. 
We then evaluate the impact of our spatial zoom-in and temporal zoom-out modules, which capture fine-grained details and long-range temporal context, respectively. Allocating all frames to a single scale (spatial zoom-in or temporal zoom-out) provides moderate gains, suggesting that relying solely on localized detail or coarse temporal coverage is insufficient for complex video reasoning. In contrast, combining both through a fixed budget split (4 spatial / 4 temporal) substantially boosts performance to 60.2 overall, demonstrating the complementary nature of the two mechanisms: spatial zoom-in improves fine-grained perception (\eg, small objects or subtle actions), while temporal zoom-out enhances global event understanding and long-range dependencies. Finally, our adaptive budget allocation achieves the best performance (60.7\% overall) and delivers the largest improvements on medium and long videos, where the optimal balance between detail and coverage varies more significantly across queries. This result highlights the importance of dynamically adjusting computational resources based on query complexity and video structure, enabling LENS to allocate frames where they are most informative.

\looseness=-1
% \noindent\textbf{Budget Allocation}. We conduct an analysis by replacing the allocator with stronger proprietary models (GPT-5-Mini and Gemini-2.5-Pro), while still using Qwen-2.5-VL for QA, which further improves Video-MME performance by +0.4\% and +0.7\%. We also note that our ablation studies show LENS still delivers significant improvements even without the controller.
\noindent\textbf{Budget Allocation}. We test whether LENS depends on the capacity of the module that allocates the frame budget between spatial zoom-in and temporal zoom-out. Replacing our default allocator with two stronger proprietary models (GPT-5-Mini and Gemini-2.5-Pro), while keeping Qwen-2.5-VL fixed as the QA backbone, yields only marginal additional gains on Video-MME (+0.4\% and +0.7\%, respectively). This indicates that the improvements stem primarily from the multi-granular zoom-in/zoom-out formulation rather than from the strength of the allocator. Consistent with this, our ablations show LENS still delivers significant gains even without the controller, demonstrating that the framework is robust to the allocation mechanism and does not rely on large proprietary models.

% When incorporating our spatio-temporal zoom-in and zoom-out modules, performance improves further. Using a fixed allocation between spatial zoom-in and temporal zoom-out leads to a strong improvement, demonstrating that jointly modeling fine-grained detail and temporal context is more effective than relying on a single scale. Finally, the proposed adaptive budget allocation achieves the best overall performance, with particularly notable gains on long videos, indicating that dynamically balancing spatial and temporal evidence is crucial for handling varying video complexities.

%% file: table/main_table.tex
\begin{table*}[t]
\centering
\small
\caption{\small \textbf{Evaluation results of VQA accuracy across different frame budgets on Video-MME, LongVideoBench (LVB), and MLVU.} Missing entries are denoted by ``–'', indicating that results are either not reported or not applicable.}
\label{tab:main-results}
\setlength{\tabcolsep}{6pt}
\renewcommand{\arraystretch}{1.07}
\resizebox{1\linewidth}{!}{%
\begin{tabular}{lcccccccc}
\toprule
\multirow{2}{*}{\textbf{Method}} 
  & \multirow{3}{*}{\textbf{Size}} 
  & \multirow{3}{*}{\textbf{\# Frames}} 
  & \multicolumn{4}{c}{\textbf{Video-MME (w.o. sub.)}} 
  & \multirow{2}{*}{\textbf{LVB}} 
  & \multirow{2}{*}{\textbf{MLVU}} \\
 & & & Short & Medium & Long & Overall & & \\
  \textit{Avg. Video Duration}& & & \textit{1.3min} & \textit{9min} & \textit{41min} & \textit{17min} & \textit{12min} & \textit{12min}\\
\midrule
\rowcolor{blue!10}\multicolumn{3}{l}{\textit{MLLMs w/ Uniform Sampling}}&&&&&&\\
Video-LLaVA~\cite{lin2024video} & 7B & 8 & 45.3 & 38.0  & 36.2  & 39.9 & 39.1 & 47.3 \\
VideoLLaMA2~\cite{cheng2024videollama} & 7B& 16 & 56.0 & 45.4  & 42.1  & 47.9  & - & - \\
LLaVA-NeXT-QW2~\cite{liu2024llavanext} & 7B& 8 & 58.0 & 47.0  & 43.4  & 49.5  & - & - \\
LongVILA~\cite{chen2025longvila} & 8B& 128 & 60.2 & 48.2  & 38.8  & 49.2  & - & - \\
LongVA~\cite{zhang2025long} & 7B & 128& 61.1 & 50.4& 46.2  & 52.6 & - & - \\
Video-XL~\cite{shu2025video}& 7B & 128/256 & 64.0 & 53.2  & 49.2  & 55.5  & - & 64.9 \\
LLaVA-OneVision~\cite{li2025llavaonevision} & 7B & * & 64.0 & 53.2  & 49.2  & 58.2  & 56.3 & 64.7 \\
LongVU~\cite{shen2025longvu} & 7B & 1fps& 64.7 & 58.2 & 59.5  & 60.9  & - & 65.4 \\
SF-LLaVA-1.5~\cite{xu2025slowfastllava} & 7B & 128& -  & -  & -  & 63.9  & 62.5 & 71.5\\
ViLAMP~\cite{cheng2025scaling} & 7B & 1fps & -  & - & 57.8  & 67.5  & 61.2 & -\\
Keye-VL-1.5~\cite{team2025kwai} & 8B & 64& 81.2	&	70.7  & 67.1  & 73.0  & 66.0 & -\\
\midrule
\rowcolor{blue!10}\multicolumn{3}{l}{\textit{Training-free Keyframe Sampling Approaches}}&&&&&&\\
\textbf{Qwen2.5-VL}~\cite{bai2025qwen25vl}     & 7B& 8 & 60.9  & 51.4  & 47.4 & 53.3  & 53.3 & 52.8 \\
+ BOLT~\cite{liu2025bolt}      & 7B& 8 & 66.0  & 54.6  & 50.4  & 57.0  & 55.6 & 59.0 \\
+ AKS~\cite{tang2025adaptive} & 7B & 8 & 62.1  & 51.4  & 48.8  & 54.1  & 52.0 &53.7 \\
+ Q-Frame~\cite{zhang2025q}& 7B  & 8 & 66.7 & 58.7  & 51.3   & 58.7   & 56.6&58.2\\
\rowcolor{gray!15}+ LENS (\textbf{Ours})        & 7B& 8 & \textbf{68.7} & \textbf{59.6}  & \textbf{53.7}  & \textbf{60.7}   & \textbf{59.3}& \textbf{63.7}\\
\midrule
\textbf{Qwen2.5-VL}~\cite{bai2025qwen25vl}    & 7B& 16 & 67.3  & 55.0  & 48.9  & 57.1  & 56.7 & 57.1 \\
+ BOLT~\cite{liu2025bolt}      & 7B& 16 & 71.7  & 57.7  & 49.7  & 59.7  & 58.0 & 64.5 \\
+ AKS~\cite{tang2025adaptive} & 7B & 16 & 66.4  & 56.9  & 48.6  & 57.3 & 55.7 & 57.8\\
+ Q-Frame~\cite{zhang2025q}& 7B & 16 & 70.0  & 59.4  & 52.6  &  60.7  & 57.1 & 61.9\\
\rowcolor{gray!15}+ LENS (\textbf{Ours})  & 7B & 16 &  \textbf{72.3}   & \textbf{62.1}  & \textbf{54.8}  & \textbf{63.1}   & \textbf{59.7} & \textbf{65.9}\\
\midrule
\textbf{Qwen2.5-VL}~\cite{bai2025qwen25vl}      & 7B& 32 & 72.6 & 59.0  & 51.8  & 61.1 & 58.4 & 59.4 \\
+ BOLT~\cite{liu2025bolt}        & 7B& 32 & 74.3 & 64.2  & 53.8  & 64.1  & 58.6 & 66.3 \\
+ AKS~\cite{tang2025adaptive} & 7B & 32& 66.4 & 56.9  & 48.6  & 57.3   & 58.3 &63.1\\
+ Q-Frame~\cite{zhang2025q} & 7B & 32 & 71.8   & 62.4 & 52.8  &  62.3   & 58.7 & 64.6 \\
\rowcolor{gray!15}+ LENS (\textbf{Ours})         & 7B& 32 & \textbf{74.6} & \textbf{70.0} & \textbf{56.7} & \textbf{67.1} & \textbf{60.6} & \textbf{66.7}\\
\midrule
\textbf{LLaVA-OneVision}~\cite{li2025llavaonevision}     & 7B& 8 & 63.9 & 51.2 & 43.9 & 53.2 & 51.8 & 59.2\\
\rowcolor{gray!15}+ LENS (\textbf{Ours})       & 7B& 8 &  \textbf{66.2} & \textbf{60.1} &  \textbf{49.7} & \textbf{58.7} & \textbf{57.5} & \textbf{66.3}\\
\midrule
\textbf{GPT-5-Mini}    & -& 8 & 74.4 & 61.1 & 58.9 & 64.8 & 55.0 & 49.1 \\
\rowcolor{gray!15}+ LENS (\textbf{Ours})        & -& 8 & \textbf{78.6} & \textbf{62.7} & \textbf{65.6} & \textbf{69.0} & \textbf{61.7} & \textbf{53.8}\\
\bottomrule
\vspace{-20pt}
\end{tabular}
}
\end{table*}

%% file: table/main_ablation.tex
\begin{table*}[t]
\centering
\small
\renewcommand{\arraystretch}{0.95}
\caption{\small \textbf{Ablation study of LENS components on Video-MME (without subtitles) using Qwen2.5-VL under an 8-frame budget}. We progressively add query-aware scoring (CLIP/BLIP), spatial zoom-in, and temporal zoom-out modules to analyze their individual and combined contributions. Note that $B_{\mathsf{spa}}$ and $B_{\mathsf{tem}}$ denote the numbers of frames allocated to spatial zoom-in and temporal zoom-out, respectively.}
\vspace{-5pt}
\label{tab:main2}
\setlength{\tabcolsep}{6pt}
\resizebox{1\linewidth}{!}{%
\begin{tabular}{lcccccc}
\toprule
\multirow{2}{*}{\textbf{Method}} &\multicolumn{2}{c}{\textbf{\# Frames}} &\multicolumn{4}{c}{\textbf{Video-MME (w.o. sub.)}}  \\
&$B_{\mathsf{spa}}$ &$B_{\mathsf{tem}}$&  Short & Medium & Long & Overall  \\
\midrule
\textbf{Qwen2.5-VL}~\cite{bai2025qwen25vl}    & - & - & 60.9  & 51.4  & 47.4 & 53.3 \\
+ CLIP Scores~\cite{radford2021learning}    & - & - & 65.8  & 53.6  & 50.4 & 56.6 \\
+ BLIP Scores~\cite{li2022blip}    & - & - & 65.9  & 54.2  & 50.8  & 57.0 \\
+ Spatial Zoom-in        & 8& -&   68.8 & 58.0 & 52.0 & 59.6 \\
+ Temporal Zoom-out       & -& 8&   68.5 & 57.5 & 51.8 & 59.2\\
+ Fixed Budgets        & 4 & 4& 68.9 & 59.0  & 52.8  & 60.2   \\
\rowcolor{gray!15}+ Adaptive Budgets       & *& *& 68.7 & 59.6  & 53.7  & 60.7   \\
\bottomrule
\vspace{-20pt}
\end{tabular}
}
\end{table*}

%% file: sec/5_conclusion.tex
\section{Conclusion}
\label{sec:conclusion}

We present LENS, a training-free, plug-and-play framework for adaptive keyframe sampling in long-form video understanding. LENS explicitly models the trade-off between fine-grained spatial evidence and long-range temporal context, allocating a limited frame budget through complementary spatial zoom-in and temporal zoom-out operations. This multi-granular strategy enables MLLMs to align visual evidence with the demands of each query, without modifying model weights or adding computational overhead.
Across diverse long-form video question answering benchmarks, LENS consistently outperforms existing keyframe sampling strategies, underscoring the value of adaptive spatio-temporal allocation for efficient video reasoning. Qualitative analyses further show that LENS yields more interpretable evidence selection, focusing on discriminative regions when fine detail is decisive and expanding temporal coverage for event-level understanding.

\section*{Acknowledgments}
\looseness=-1
This work has been funded in part by the ARL award W911QX-24-F-0049, DARPA award FA8750-23-2-1015, ONR award N00014-23-1-2840, ONR MURI grant N00014-25-1-2116, and USDA-NIFA award 2021-67021-35329.

%% file: sec/X_appendix.tex
\renewcommand{\thesection}{\Alph{section}}
\renewcommand{\theHsection}{\Alph{section}}
\renewcommand\thefigure{\Alph{section}\arabic{figure}} 
\renewcommand\thetable{\Alph{section}\arabic{table}}  
\setcounter{section}{0}
\setcounter{figure}{0} 
\setcounter{table}{0}

{
\newpage
    \centering
    \Large
    \textbf{LENS: Adaptive Spatio-Temporal Zooming for Keyframe Sampling in Long-Form Videos}\\
    \vspace{0.5em}Appendix\\
    \vspace{1.0em}
}
In the appendix, we provide additional experimental results and implementation details of our LENS framework.

\section{More Experimental Results}
\label{appendix:exp}
\subsection{More Ablation Studies}
The ablation results in Table~\ref{tab:main1} examine the impact of the key hyperparameters in the temporal zoom-out branch, including the Gaussian kernel bandwidth $\sigma$, the graph propagation coefficient $\alpha$, and the hyperframe size $\tau$. Overall, the performance remains relatively stable across a wide range of settings, suggesting that the proposed framework is robust to moderate hyperparameter variations. In particular, setting $\sigma=0.3$ yields the best overall performance, while both smaller ($\sigma=0.1$) and larger ($\sigma=1$) bandwidths slightly degrade accuracy, indicating that a moderate similarity scale provides the most reliable graph affinity structure. The propagation coefficient $\alpha$ also plays an important role: removing graph propagation ($\alpha=0$) slightly reduces overall accuracy, while overly strong propagation ($\alpha=1$) leads to the largest drop, suggesting that balancing the initial query-frame similarity with the global video structure is important for stable reasoning. Finally, the hyperframe size $\tau$ controls the temporal context aggregated around each selected frame. Using $\tau=4$ achieves the best overall result, while smaller ($\tau=1$) or larger ($\tau=9$) windows slightly hurt performance, indicating that moderate temporal aggregation provides sufficient contextual cues without excessively sacrificing spatial detail. These results collectively demonstrate that LENS benefits from a balanced design that jointly considers local evidence and global temporal structure.
\vspace{-10pt}
\begin{table}[H]
\centering
\small
\caption{\small \textbf{Ablation study of key hyperparameters in the temporal zoom-out branch on Video-MME (w.o. subtitles).} 
We vary the Gaussian kernel bandwidth $\sigma$ used in the graph affinity matrix, the propagation coefficient $\alpha$ controlling the strength of graph message passing, and the hyperframe size $\tau$ that determines the temporal window aggregated for each selected frame. }
\vspace{-5pt}
\label{tab:main1}
\setlength{\tabcolsep}{6pt}
\resizebox{0.8\linewidth}{!}{%
\begin{tabular}{clcccc}
\toprule
\multirow{2}{*}{\#}&\multirow{2}{*}{\textbf{Settings}}  &\multicolumn{4}{c}{\textbf{Video-MME (w.o. sub.)}}  \\
&&Short & Medium & Long & Overall  \\
\midrule
1&$\sigma = 0.3$, $\alpha = 0.8$, $\tau = 4$     & 68.7 & 59.6  & 53.7  & 60.7 \\
2&$\sigma = 0.1$, $\alpha = 0.8$, $\tau = 4$     & 68.5 & 59.3  & 53.4  & 60.4 \\
3&$\sigma = 1$, $\alpha = 0.8$, $\tau = 4$     & 68.5 & 59.1  & 53.0  & 60.2 \\
4&$\sigma = 0.3$, $\alpha = 0$, $\tau = 4$     & 69.1 & 58.9  & 52.7  & 60.2 \\
5&$\sigma = 0.3$, $\alpha = 1$, $\tau = 4$     & 67.8 & 57.5 & 51.8 & 59.0\\
6&$\sigma = 0.3$, $\alpha = 0.8$, $\tau = 1$     & 69.0 & 57.9 & 52.3 & 59.7\\
7&$\sigma = 0.3$, $\alpha = 0.8$, $\tau = 9$     & 68.2 & 59.3 & 53.5 & 60.3 \\
\bottomrule
\vspace{-20pt}
\end{tabular}
}
\end{table}

\subsection{More Benchmarks}
We further evaluate on EgoSchema and NExT-QA under different frame budgets; results are shown below. LENS demonstrates consistent performance gains across benchmarks.

\begin{table}[H]
\vspace{-12pt}
\centering
\small
\setlength{\tabcolsep}{6pt}
\begin{minipage}{0.49\linewidth}
\centering
\setlength{\aboverulesep}{0pt}
\setlength{\belowrulesep}{2pt}
\begin{tabular}{lccc}
% \toprule
\textbf{EgoSchema} & 8 & 16 & 32 \\
\midrule
Qwen-2.5-VL & 59.2 & 61.2 & 62.6 \\
\rowcolor{gray!15}
\textbf{+ LENS}         & \textbf{60.6} & \textbf{62.7} & \textbf{66.1} \\
% \bottomrule
\end{tabular}
\end{minipage}
\hfill
\begin{minipage}{0.49\linewidth}
\centering
\setlength{\aboverulesep}{0pt}
\setlength{\belowrulesep}{2pt}
\begin{tabular}{lccc}
% \toprule
\textbf{NExT-QA} & 8 & 16 & 32 \\
\midrule
Qwen-2.5-VL & 74.7 & 78.4 & 79.1 \\
\rowcolor{gray!15}\textbf{+ LENS}         & \textbf{76.6} & \textbf{79.4} & \textbf{80.3} \\
% \bottomrule
\end{tabular}
\end{minipage}
\label{tab:main}
\end{table}
\vspace{-15pt}

\subsection{MLVU Results}
We also provide category-wise results on MLVU with 8 frames below. Our LENS consistently improves performance across categories. 

\vspace{-12pt}
\begin{table}[H]
\centering
\footnotesize
\setlength{\tabcolsep}{2pt}
\renewcommand{\arraystretch}{0.85}
\resizebox{\linewidth}{!}{
\setlength{\aboverulesep}{0pt}
\setlength{\belowrulesep}{2pt}
\begin{tabular}{lcccccccc}
% \toprule
\textbf{Method} 
& TR & AR 
& NQA & ER & PQA 
& AO & AC 
& Avg. \\
\midrule
Qwen-2.5-VL 
& 80.8 & 60.6
& 55.6 & 53.3 & 52.9 
& 43.2 & 19.0 
& 52.8 \\
\rowcolor{gray!15}
\textbf{+ LENS} 
% & \textbf{87.5} & \textbf{66.5} 
% & \textbf{74.4} & \textbf{59.5} & \textbf{69.2} 
% & \textbf{53.3} & \textbf{41.3} 
% & \textbf{65.9} \\
& \textbf{85.9} & \textbf{61.0} 
& \textbf{71.8} & \textbf{57.3} & \textbf{67.5} 
& \textbf{52.1} & \textbf{40.9} 
& \textbf{63.7} \\
% \bottomrule
\end{tabular}
}
\vspace{-15pt}
\label{tab:category_results}
\end{table}

\section{More Implementation Details}
\label{appendix:implementation}
\subsection{Full Implementation Details}
 For all evaluations, we follow the default query formats and prompts provided by the respective benchmarks to ensure consistent evaluation. Following prior work~\cite{tang2025adaptive,zhang2025q}, we evaluate under frame budgets of 8, 16, and 32 frames per video. Frame–query similarity scores used for initial keyframe selection are computed using a pre-trained CLIP ViT-L/14 model. For spatial zoom-in, attention maps derived from CLIP are used to generate query-guided visual prompts that highlight relevant regions without altering the original frame content.

For the temporal zoom-out module, we construct a video graph where each node corresponds to a frame representation and edges encode visual similarity between frames. Pairwise similarities are computed using CLIP image–image cosine similarity, and the affinity matrix is defined using a Gaussian kernel. Message passing is performed with the update rule described in Sec.~\ref{sec:method} until convergence. The hyperparameters are set to $\sigma=0.3$ for the Gaussian kernel bandwidth, $\alpha=0.8$ for the propagation coefficient, and $\tau=4$ for the hyperframe size, which determines the number of neighboring frames concatenated to form each hyperframe. These values are selected based on ablation studies reported in the Appendix.

To avoid redundancy between the two branches, frames selected by the spatial zoom-in module are excluded from the candidate pool of the temporal zoom-out module. The adaptive budget allocation ratio is predicted using the MLLM controller described in Sec.~\ref{sec:method}, which analyzes the input question and outputs the proportion of frames assigned to each branch. Finally, the selected frames from both branches are merged and ordered according to their original temporal indices before being fed into the downstream MLLM for video question answering.

All experiments are conducted on a server equipped with 8$\times$48GB NVIDIA RTX 6000 Ada GPUs. Unless otherwise specified, inference is performed with default decoding parameters of the respective MLLMs.

\subsection{Prompts}
The prompt used to guide this allocation decision is provided as follows.
\begin{tcblisting}{
colback=blue!5!white,
colframe=blue!75!black,
breakable,
title={\textsc{Prompt Used for Budget Allocation}},
listing only
}
ALLOCATION_PROMPT = """
You are a Video Resource Controller for video QA.

Output r in [0,1]: fraction of frames for SPATIAL ZOOM-IN.
(1-r) is for TEMPORAL ZOOM-OUT.

Use SPATIAL when: read text/numbers, small objects, attributes (color/logo), fine details.
Use TEMPORAL when: main idea/summary, events, before/after, sequence, changes over time, count occurrences.

Anchor:
detail -> r=0.7-0.95
temporal -> r=0.05-0.30
mixed -> r=0.40-0.60

Examples (Query -> r):
"What is the main idea of the video?" -> 0.20
"What happens at the end of the video?" -> 0.25
"How many times does the person jump?" -> 0.25
"Which smartphone is advertised on the laptop screen?" -> 0.83
"What word is written on the sign?" -> 0.85
"What color is the car?" -> 0.75
"Which element doesn't show up in the video?" -> 0.35
"What object does the person pick up and when?" -> 0.50
"What is the sequence of steps introduced in the video?" -> 0.15

Now answer for:
Query: "{USER_QUERY}"

OUTPUT ONLY r (one number). No other text.
"""
\end{tcblisting}

\section{Limitations and Broader Impacts}
\noindent\textbf{Limitations}. Although LENS improves keyframe selection for long-form video understanding, we identify two potential limitations. First, the proposed framework relies on pre-trained vision–language models such as CLIP to estimate frame–query relevance. Consequently, the quality of the selected frames is inherently bounded by the representation capability of these models, which may struggle with highly abstract queries or domain-specific content. Second, while the temporal zoom-out module captures global video structure through graph-based reasoning, the constructed graph is still based on pairwise visual similarity and may not fully capture complex causal relationships across distant events.

\noindent\textbf{Broader Impacts}. Efficient long-form video understanding has the potential to improve numerous real-world applications, including video search, multimedia analysis, educational content indexing, and assistive technologies. By enabling models to reason over long videos using limited visual tokens, LENS may help reduce computational costs and make large-scale video understanding systems more accessible. However, improved video analysis capabilities could potentially be misused for large-scale surveillance or automated monitoring without appropriate safeguards. Responsible deployment should therefore consider transparency, privacy protection, and fairness considerations when applying such technologies in real-world settings.